\documentclass{article}
\usepackage{spconf,amsmath,graphicx}
\usepackage{float,booktabs}
\usepackage{enumitem}
\usepackage[noend]{algpseudocode}
\usepackage{subfigure}
\usepackage[linesnumbered,ruled,vlined, noend]{algorithm2e}

\usepackage{xcolor}

\title{D-CBRS: Accounting For Intra-Class Diversity in Continual Learning}
\name{Yasin Findik and Farhad Pourkamali-Anaraki\thanks{© 20XX IEEE. Personal use of this material is permitted. Permission from IEEE must be obtained for all other uses, in any current or future media, including reprinting/republishing this material for advertising or promotional purposes, creating new collective works, for resale or redistribution to servers or lists, or reuse of any copyrighted component of this work in other works.}}
\address{Department of Computer Science, University of
Massachusetts Lowell, MA, USA}
\begin{document}
%
\maketitle

\begin{abstract}
Continual learning -- accumulating knowledge from a sequence of learning experiences -- is an important yet challenging problem. In this paradigm, the model's performance for previously encountered instances may substantially drop as additional data are seen. When dealing with class-imbalanced data, forgetting is further exacerbated. Prior work has proposed replay-based approaches which aim at reducing forgetting by intelligently storing instances for future replay. Although Class-Balancing Reservoir Sampling (CBRS) has been successful in dealing with imbalanced data, the intra-class diversity has not been accounted for, implicitly assuming that each instance of a class is equally informative. We present Diverse-CBRS (D-CBRS), an algorithm that allows us to consider within class diversity when storing instances in the memory. Our results show that D-CBRS outperforms state-of-the-art memory management continual learning algorithms on data sets with considerable intra-class diversity.
\end{abstract}
\section{Introduction}
\label{sec:intro}

Continual learning examines the problem of learning from data streams by accommodating new knowledge while retaining previously learned experiences \cite{chen}. In such an incremental learning setting, a model is supposed to learn new tasks, domains, or classes with incoming data.  For example, in class-incremental learning, analogous to human experience, incoming streams continuously introduce new classes (i.e., knowledge) that are expected to be learned~\cite{vandeven, Linda}.

Yet, unsurprisingly, continual learning leads to forgetting: as we encounter new classes, the model's performance may substantially degrade regarding previously acquired patterns
\cite{FRENCH, McCloskey, goodfellow}. 
Current approaches that seek to reduce catastrophic forgetting for class-incremental learning primarily include \textit{rehearsal-based approaches} \cite{Chaudhry,Rahaf} which store samples that belong to previously observed classes and replay them while receiving a data stream 
and  \textit{regularization-based approaches} \cite{Li,kirkpatrick} which introduce regularization terms to loss functions to emphasize existing classes. 

Most rehearsal-based methods use either ring buffer \cite{Chaudhry} or reservoir sampling \cite{Vitter} as memory management. Because these methods rely on storing and replaying previously seen data, when dealing with imbalanced datasets, the replay is overwhelmed by the ``overrepresented'' classes. Thus, it exacerbates catastrophic forgetting even further ~\cite{wu}.

Recently, a novel approach was proposed for managing memory in the presence of imbalanced data, named as Class-Balancing Reservoir Sampling (CBRS) \cite{Chrysakis}. In a nutshell, CBRS keeps the memory balanced in terms of the number of instances per class. In this way, CBRS alleviates the possible forgetting issues regarding underrepresented classes.

We posit that catastrophic forgetting, in addition to class imbalance, can also be substantially exacerbated by overlooking the intra-class diversity. While CBRS keeps the classes in memory balanced, it does not account for the intra-class diversity because it treats every data instance as equally informative or representative of the respective class. Hence, building upon CBRS, our motivation lies in finding a new way to incorporate the intra-class diversity when storing incoming data in the memory. Our proposed method, Diverse Class-Balancing Reservoir Sampling (D-CBRS), in addition to maintaining the class balance, also maximizes the intra-class diversity by storing more diverse instances in the memory.
We conduct experiments to compare D-CBRS with state-of-the-art memory management approaches on two different datasets. In our experimental setup, the results show that D-CBRS outperforms the other approaches. We further discuss the limitations of our approach, and propose future improvements. Our findings indicate that accounting for the intra-class diversity presents a new research direction in the continual learning space.

\begin{figure*}
    \centering
    \begin{subfigure}{}
      \centering
      \includegraphics[width=0.23\linewidth]{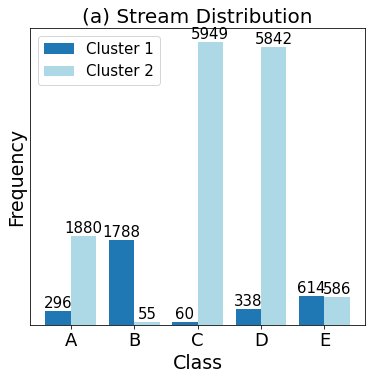}
    \end{subfigure}
    \begin{subfigure}{}
      \centering
      \includegraphics[width=0.23\linewidth]{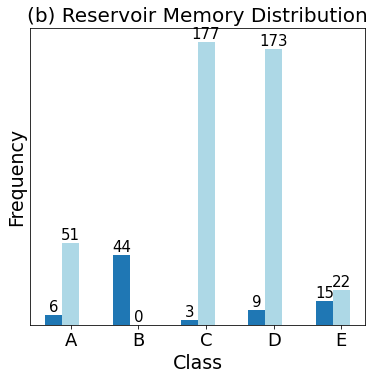}
    \end{subfigure} 
    \begin{subfigure}{}
      \centering
      \includegraphics[width=0.23\linewidth]{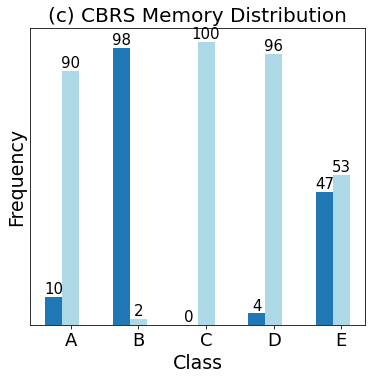}
    \end{subfigure}
    \begin{subfigure}{}
      \centering
      \includegraphics[width=0.23\linewidth]{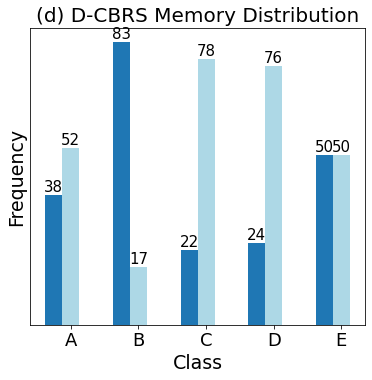}
    \end{subfigure}
    \caption{(a) Sample of imbalanced and intra-class diverse incoming data distribution with 5 classes. Cluster 1 and cluster 2 indicate the intra-class diversity for each class (i.e., each class consists of two different clusters of instances). Reservoir (b) and CBRS (c) store substantially less instances of the minority clusters in memory compared to our proposed D-CBRS (d). } 
    \label{fig:figure}
\end{figure*}

\section{PROPOSED METHOD AND RELATION TO PRIOR WORK}

The proposed method builds upon CBRS.
For this reason, we first describe CBRS, its notation, and how it achieves class balance. Subsequently, we delve into D-CBRS, show where it differs from CBRS, and how it preserves the within-class diversity. Finally, we delineate the steps for training a model with replay under class incremental learning.

\subsection{CBRS} 
\label{CBRS}
CBRS is an algorithm that determines which data instances will be stored in the memory for future replay. While storing the data samples, it also seeks to keep the classes in the memory as balanced as possible.   

\textbf{Notation. }The size of a given memory is noted with \textit{m}. A \textit{filled} memory refers to all \textit{m} spots being occupied with data instances. Once the memory is filled, CBRS labels the largest class in the memory (in terms of the number of instances) as a \textit{full} class. Once a class is labeled as \textit{full}, it remains so. 

\textbf{Algorithm. }CBRS is characterized by two distinct stages. The first stage lasts until memory is \textit{filled}. During the first stage, all the incoming instances are stored in the memory. In the second stage, which starts after the memory is \textit{filled}, when a data instance $(x_i, y_i)$ is encountered, CBRS first checks whether the instance is a member of a \textit{full} class. If so, the instance will be replaced by one of the other instances which belongs to class $y_i$ with probability $\frac{m_c}{n_c}$, where $m_c$ is the number of instances of class $y_i$ in memory, $n_c$ is the total number of instances of class $y_i$ received so far. If the instance does not fall into \textit{full} classes, the new instance is stored in the memory by removing randomly an instance from a \textit{full} class.

\subsection{D-CBRS} 
As mentioned earlier, CBRS does not account for the within-class diversity as it treats every data instance as equally relevant and informative of the class. To address this problem, our proposed method aims to incorporate information regarding diversity within classes. The main idea behind D-CRBS is that it clusters instances within each class and keeps track of the number of instances of each cluster within each class. In this way, it ensures that instances from smaller clusters, which often hold rich information for inter-class discriminatory purposes, will be stored. Consequently, the proposed approach alleviates the forgetting problem since it will keep such rare instances in the memory. Figure \ref{fig:figure} illustrates this point.

\textbf{Algorithm.} Similar to CBRS, initially D-CBRS stores all instances until the memory is \textit{filled}. The difference between CBRS and D-CBRS starts in the second stage where the removal of instances from the memory occurs in CBRS. When the memory is filled, an incoming instance $(x_i, y_i)$ can either belong to a \textit{full} class or not.
For the first case, if $y_i$ is not one of the \textit{full} classes, D-CBRS stores the sample by replacing with a randomly picked instance from the largest cluster of any of the full classes in the memory. In the second case, when $y_i$ is a member of a \textit{full} class, D-CBRS first checks if the incoming instance belongs to the largest cluster of that class $y_i$. If so, the new instance takes a randomly chosen instance's place in that cluster with probability $\frac{m_c}{n_c}$, where $m_c$ is the number of instances of the cluster in memory and $n_c$ is the total number of instances of the cluster encountered up to that point. Otherwise, it will be replaced by a randomly selected instance from the largest cluster of class $y_i$. The pseudo-code for D-CBRS can be found in Algorithm ~\ref{alg:D-CBRS}. Note that in our experiments, we used the K-means clustering algorithm \cite{Lloyd, pourkamali2017preconditioned} since its implementation is straightforward and it works reasonably well for MNIST and F-MNIST datasets that we will use later. However, D-CBRS can work with other clustering algorithms, including spectral clustering \cite{pourkamali2020scalable}.

\renewcommand{\algorithmicrequire}{\textbf{Input:}}
\renewcommand{\algorithmicensure}{\textbf{Output:}}

\begin{algorithm}[t]
    \caption{D-CBRS}
    \label{alg:D-CBRS}
    \SetAlgoLined
    \SetKwInOut{Input}{input}
    \SetKwInOut{Output}{output}
    \DontPrintSemicolon
    
    \Input{data stream arriving sequentially ${(x_i, y_i)}^n_{i=1}$}
    \Output{updated memory}
    
    \For{$i=1, \ldots, n$}{
        \eIf{memory is \textbf{not} filled}{
            store ($x_i, y_i$)
        }{
            \eIf{$y_i$ is \textbf{not} a full class}{
                cluster a randomly picked full class 
                choose an instance randomly from largest cluster of the full class and replace by ($x_i, y_i$)
            }{
                cluster full class $y_i$
                
                $(x_p, y_p) \gets \text{pick an instance randomly}$
                
                \text{from largest cluster of the full class}
                    
                \eIf{$x_i$ belongs to largest cluster}{
                    $m_c \gets \text{number of currently}$
                    
                    \text{stored instances belongs to }
                    
                    \text{the largest cluster of $y_i$}
                    
                    $n_c \gets \text{number of instances}$
                    
                    \text{seen so far belongs to}
                    
                    \text{the largest cluster of $y_i$}
                    
                    $u \gets {Uniform(0, 1)}$
                    
                    \If{$u \leq m_c/n_c$}{
                        replace ($x_p, y_p$) by ($x_i, y_i$)
                        
                    }
                }{
                    replace ($x_p, y_p$) by ($x_i, y_i$)
                }
            }
        }
    }
\end{algorithm}



\subsection{Training with replay} 
In class-incremental scenarios, streams of instances from each class arrive sequentially and each class appears at a time only. 
Assume that $S$ is the set of streams $\{s_1, s_2, ... s_n\}$, where $s_i$ consists of class $i$ instances. The model will be trained from $s_1$ to $s_n$. The model is then expected to remember all classes. 

In rehearsal approaches, it is assumed that there is a limited memory space to store samples during training and the model is allowed to remember previously seen classes, along with replaying the memory for subsequent training steps. 

For the training phase, we follow the same steps described in \cite{Chaudhry,Aljundi}, where CBRS originated from. Implementation of the algorithm is as follows: each incoming stream is split to batches. Then each batch is concatenated with a sample batch stemming from the memory. The obtained batch is used to train the model. We repeat the concatenation step $n_b$ times, for each of which we concatenate the same incoming batch with a different sampled batch from the memory.
Thus, the model is updated $n_b$ times when a new batch stream is received. This type of replay reduces catastrophic forgetting further because the model sees both the new batch multiple times and random batches from the memory in each update step.
Lastly, the memory will be updated according to incoming stream instances. 

Once the training phase is completed, we evaluate the performance of the model on an unseen test set, which consists of all the classes of the original dataset.

\section{EXPERIMENTS}
\label{sec:typestyle}

In this section, we describe our experimental setup and show the results of our algorithm and its comparison with existing memory management algorithms for rehearsal-based continual learning scenarios. Following CBRS, we use MNIST \cite{deng} and Fashion-MNIST \cite{xiao} datasets in this section.

\subsection{Experimental Setup}

\subsubsection{Memory Management Approaches} 

We compare D-CBRS with standard memory management algorithms for rehearsal-based approaches including reservoir sampling (RS) and class-balancing reservoir sampling (CBRS). Since CBRS outperforms Gradient-Space Sampling (GSS)~\cite{Aljundi}, another memory management approach, we do not include GSS in our experiments. 
Although RS is also surpassed by CBRS, we include it in our comparisons as CBRS has evolved from it and it provides a useful baseline. Figure \ref{fig:figure} shows the sample distribution and memory allocation for each of the memory management approaches. 


\textbf{RS} takes an input stream of data, then returns a random subset of instances from it. Each sample is stored with probability $\frac{size}{n}$, where \textit{size} is the memory size and \textit{n} is the total number of observed samples so far.

\textbf{CBRS} follows a similar strategy as that of RS while keeping classes stored in the memory balanced. Detail explanation can be found in Section \ref{CBRS}. 

\subsubsection{Simulating Class Imbalance}
\label{class_imbalance}
We followed the technique described in the CBRS paper~\cite{Chrysakis} to generate class imbalance scenarios. As suggested, we start by defining a set of retention factors, i.e., what percentage of instances for each class from the original dataset will be used in the training phase. The used retention factor set for our experiments is as follows:

\begin{equation}
r = \{0.01, 0.05, 0.1, 0.3, 1\}.
\end{equation}

Second, we randomly choose a retention factor from $r$ for each class in the dataset without replacement. If the number of classes is greater than the defined retention factor set size, once all the instances of the retention set are used we reinitialize our retention factor set as $r$. We ensure that identical class-imbalanced datasets are used for evaluating all memory management methods.

\subsubsection{Simulating Intra-class Diversity}
\label{intra_class_diversity}
To see the effect of the intra-class diversity on the memory management methods, we simulate the diversity by randomly grouping together original classes of the original dataset to form more diverse classes. For example, given a dataset with $10$ classes and our target number of classes $5$, we create a new class by combining randomly two classes and updating the instances of those classes by labeling the new class. When we do this for all classes, we will have 5 new classes and each one of them will consist of instances that belong to two old classes. In this way, we make sure that there will be a sufficient level of diversity within the newly-created classes.

Each independent run may produce a different dataset depending on which old classes are combined together. Once sampled, all the methods are evaluated on the same dataset for a single run.




 
\subsubsection{Models and Hyperparameters}
Since the two studied datasets are simple enough not to require a more complex model, and also to compare the results better with CBRS, we used the exact same model: a multi-layer perceptron (MLP) with 2 hidden layers, 250 neurons per hidden layer, and the ReLU activation function.



We use \textit{Adam} \cite{diederik} optimizer with a learning rate
of $0.001$ and \textit{cross-entropy} loss.
Following CBRS, we assign batch size as $10$ and the number of steps per batch as $5$. We fix memory size as $m=500$ samples in our experiments. 

\section{Results} 

In this section, we present the results of our approach in three different setups. 

\subsection{Base Case -- Class Imbalance Only}
In the first setup, we simulated only class imbalance in the dataset as described in Section~\ref{class_imbalance}. Since the two datasets are relatively simple to learn and do not have substantial intra-class diversity, this served as a check on whether D-CBRS achieves the same accuracy as CBRS when only class imbalance exists in the dataset. As expected, our results from 5 different random runs show that D-CBRS achieves approximately similar accuracy as CBRS on both MNIST and F-MNIST data sets (See Table~\ref{Experiment 1} (a)).

\subsection{Omniscient Case -- Class Imbalance \& Intra-Class Diversity With Labels}

In the second case, we ran the algorithms on imbalanced and diverse datasets (simulated as described in Section \ref{intra_class_diversity}).
To understand the maximum accuracy that D-CBRS would achieve if it could perfectly predict diversity of data instances, we allowed D-CBRS access to actual class labels. For example, in the MNIST case, if class $A$ was new class from digits 2 and 9, D-CBRS would be aware of the labels of the two composing classes. In this way, we were able to disentangle the accuracy of the clustering algorithm from the accuracy of our proposed method. The results (Table \ref{Experiment 1} (b)) show (1) how CBRS suffers when intra-class diversity is present and (2) the substantial improvement in terms of accuracy that D-CBRS would be capable of if it could query an oracle regarding intra-class diversity. In other words, this experiment shows the significance of leveraging powerful clustering algorithms in our framework. 

\begin{table}
\begin{tabular}{@{}ccccc@{}}
\toprule
Method    & \multicolumn{4}{c}{Dataset}                                                      \\ \midrule
          & \multicolumn{2}{c|}{Imbalanced (a)}      & \multicolumn{2}{c}{Imbalanced \& Div. (b)} \\ \cmidrule(l){2-5} 
          & MNIST & \multicolumn{1}{c|}{F-MNIST} & MNIST              & F-MNIST              \\
Reservoir &    66.9 ± 3.8   &  \multicolumn{1}{c|}{62.3 ± 2.2}      &            66.3 ± 1.2        &       64.7 ± 2.2              \\
CBRS      &  82.8 ± 3.5      & \multicolumn{1}{c|}{75.0 ± 2.1}        &        76.8 ± 2.3            &        70.3 ± 1.9              \\
D-CBRS    &   80.6 ± 3.2    & \multicolumn{1}{c|}{73.1 ± 3.1}        &          84.6 ± 0.4          &     76.9 ± 1.2                 \\ \bottomrule
\end{tabular}
\caption{
Results of 5 streams with 95\% confidence interval of the accuracy on the test set for (a) \em Base Case \em and (b) \em Omniscient Case \em with \em 5 \em total resulting merged classes.}
\label{Experiment 1}
\end{table}

\subsection{Realistic Case -- Class Imbalance \& Intra-Class Diversity Without Labels}

Realistically, D-CBRS has to rely on a diversity scoring algorithm, such as a clustering one, since it cannot have access to a ``diversity scoring oracle''. In this case, the performance of D-CBRS is dependent on the performance of the underlying clustering algorithm (e.g., K-means in our experiments). Our results show that D-CBRS outperforms CBRS even with a relatively simple algorithm as K-means; see Table \ref{Experiment 2}.
While D-CBRS outperforms CBRS when intra-class diversity is present, it should be noted that it does so at a computational cost -- the clustering algorithm's running complexity. We show intra-class diversity to be an important aspect of continual learning, and suggest that future work should focus on developing algorithms that improve both the performance of the diversity scoring and the computational complexity of D-CBRS.





\begin{table}
\begin{tabular}{@{}lcccc@{}}
\toprule
\multicolumn{1}{c}{Method} & \multicolumn{4}{c}{Imbalanced \& Diverse Dataset}                    \\ \midrule
\multicolumn{1}{c}{}       & \multicolumn{2}{c|}{5 Classes}       & \multicolumn{2}{c}{3 Classes} \\ \cmidrule(l){2-5} 
\multicolumn{1}{c}{}       & MNIST & \multicolumn{1}{c|}{F-MNIST} & MNIST        & F-MNIST        \\
Reservoir                  & 66.3 ± 1.6      & \multicolumn{1}{c|}{65.5 ± 3.9}        &      73.7 ± 3.8       &       75.8 ± 5.5         \\
CBRS                  &    75.2 ± 3.0        & \multicolumn{1}{c|}{70.9 ± 3.7}        &     72.7 ± 5.2         &      72.8 ± 5.4          \\
D-CBRS                     &   79.7 ± 1.1    & \multicolumn{1}{c|}{72.9 ± 3.0}        &    80.6 ± 3.5          &        78.0 ± 4.5        \\ \bottomrule
\end{tabular}
\caption{Results of 5 runs with 95\% confidence interval of the accuracy on the test set for \em Realistic Case \em with (a) 5 and (b) 3 total resulting merged classes.}
\label{Experiment 2}
\end{table}


\section{Conclusion}

Previous work in continual learning has proposed algorithms for dealing with imbalanced data distributions. In this work, we showed that in addition to class imbalance, intra-class diversity is also a factor that substantially affects performance and the catastrophic forgetting phenomenon. Thus, it needs to be accounted for in the design of memory management algorithms for class-incremental learning settings. We proposed D-CBRS as an approach to consider the intra-class diversity and class imbalance in such settings. Our experimental results showed that D-CBRS outperforms previous approaches  when the distribution of the data is imbalanced and diverse. Future research directions involve improving efficiency and accuracy of D-CBRS, exploring more complex data sets, and understanding tradeoffs for varying memory sizes. 




\bibliographystyle{IEEEbib}
\bibliography{refs}

\end{document}